\documentclass{article} 
\usepackage{iclr2023_conference_tinypaper,times}

\usepackage{amsmath,amsfonts,bm}

\def\eqref#1{equation~\ref{#1}}

\def\1{\bm{1}}

\DeclareMathAlphabet{\mathsfit}{\encodingdefault}{\sfdefault}{m}{sl}
\SetMathAlphabet{\mathsfit}{bold}{\encodingdefault}{\sfdefault}{bx}{n}

\usepackage{hyperref}
\usepackage{url}
\usepackage{graphicx}
\usepackage{array}

\title{Zero-shot generalization across\\ architectures for visual classification}

\author{Evan Gerritz$^*$, Luciano Dyballa$^*$,\ \ Steven W. Zucker\\
Dept. of Computer Science, Yale University, New Haven, CT, USA \qquad {\tiny $^*$ Equal contribution}
}

\iclrfinalcopy 
\begin{document}

\maketitle
\begin{abstract}
Generalization to unseen data is a key desideratum for deep networks, but its relation to classification accuracy is unclear.
Using a minimalist vision dataset and a measure of generalizability, we show that popular networks, from deep convolutional networks (CNNs) to transformers, vary in their power to extrapolate to unseen classes both across layers and across architectures.
Accuracy is not a good predictor of generalizability, and generalization varies non-monotonically with layer depth.
Our code is available at \href{https://github.com/dyballa/generalization/tree/ICLR2024TinyPaper}{{\fontfamily{lmss}\selectfont github.com/dyballa/generalization}}.

\end{abstract}

\section{Introduction}



In deep learning for classification, generalization is typically considered in the context of training versus test sets \citep{zhang2021understanding}, where both contain examples from the same set of classes, and is measured via test set accuracy.
Our goal is to investigate the capacity of a network to generalize its classification power to similar classes absent in the training set (unseen). 
We work within the domain of Chinese calligraphy. It consists exclusively of ink on paper, making it a particularly challenging instance of the artist recognition problem and also related to the classic problem of separating style from content \citep{tenenbaum2000separating}.
It is easy to show that many networks can classify characters by artist, as expected, from the training set \citep{vgg_dataset}, but generalization remains largely unexamined. 
We adopt an embedding-based zero-shot paradigm, commonly applied to text-to-image models \citep{xian2018zero}, but here applied to pure vision models, i.e., with no external semantic information.  In the end, (\textit{i}) our approach leads to a  method to quantify generalization in this minimalist domain; (\textit{ii}) accuracy of classification is not a good predictor of classification generalizability, i.e., the ability to generalize to out-of-sample (unseen) artists; and (\textit{iii}) different architectures yield surprisingly different latent representations.

\section{Methods \& results}\label{lbl:methods}

We fine-tuned an ensemble of pretrained networks\footnote{ResNet50 \citep{resnet}, ViT-base \citep{vit}, Swin Transformer \citep{swin}, 
Pyramid ViT (PViT) \citep{pvt}, CvT-21 \citep{cvt}, PoolFormer-S12 \citep{poolformer}, ConvNeXt V2 \citep{convnext2}} to classify artists from the calligraphy dataset of \citet{dataset}.  Using the intuition that learned classes should appear more clustered in the space of intermediate representations, we develop a proxy for generalization based on the degree of separability that unseen classes achieve in the latent embeddings. To quantify zero-shot classification, we define a \emph{generalization index}, $g$, for a given network, based on the normalized mutual information \citep{danon2005comparing} between a k-means cluster assignment computed on the embedding of unseen examples and the ground truth: $g = \max_i \left\{\mathrm{NMI}\left ( \mathcal{C}^{i}_{\mathrm{unseen}}, \mathcal{C}^{\star}\right)\right\}$
where $i$ indexes the intermediate layer outputs, and $\mathcal{C}_{\mathrm{unseen}}$ and $\mathcal{C}^\star$ denote cluster assignments for unseen examples and ground truth, respectively. To establish a baseline for generalization, we also computed $g$ for the classes the network was trained on ($g_{\text{seen}}$) (Table \ref{table:network_comparison}). To validate our choice of metric, we tested an additional metric based on $k$-nearest neighbors (see Appendix~\ref{lbl:app-metric}); finally, to ensure our findings were consistent with a standard dataset, we repeated our tests using CIFAR-100 \citep{Krizhevsky09learningmultiple} (see Appendix~\ref{lbl:app-cifar}).

\begin{figure}[ht]
    \begin{center}
        \includegraphics[width=1\textwidth]{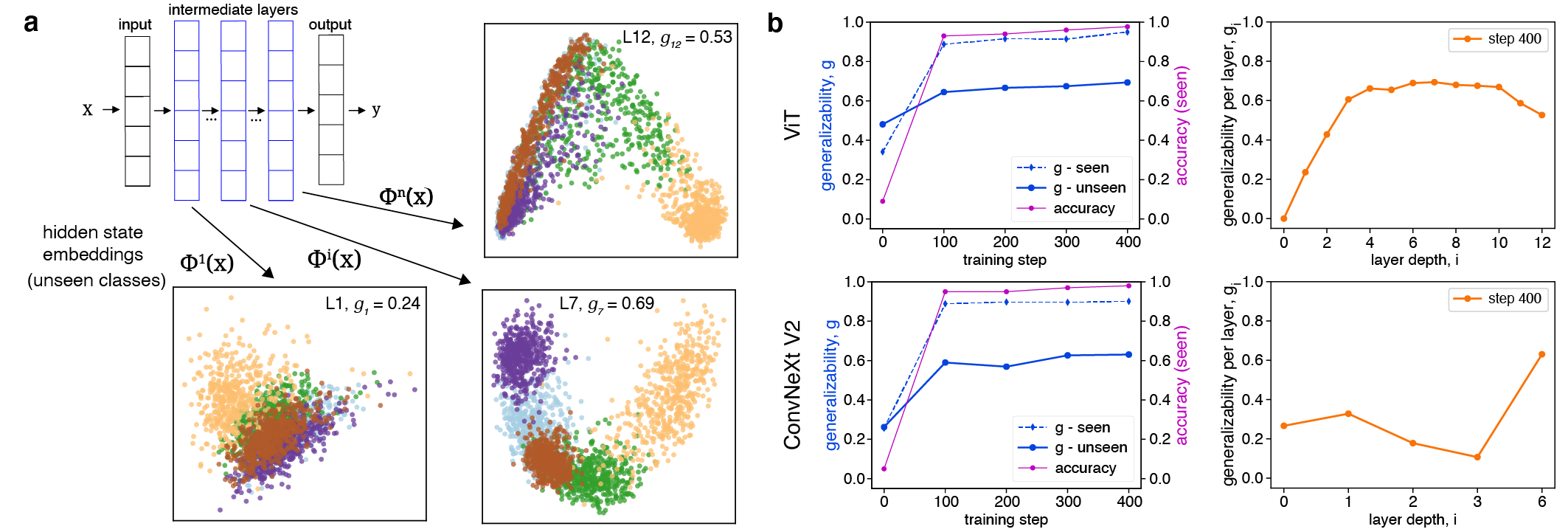}       
    \end{center}
    \caption{{\bf Generalizability and accuracy are loosely coupled.} (\textbf{a}) An overview of our method of assessing generalizability through out-of-sample embeddings using intermediate layers, visualized using PCA. Embeddings from different hidden states of the Vision Transformer (ViT) produce widely varying results. Color labels indicate ground truth: clustered unseen classes indicate better generalization ($g$). (\textbf{b}) Results for ViT (top) and ConvNeXtV2 (bottom).  Across epochs, test-set accuracy monotonically increases while generalizability may plateau or even decrease (left). Across layers, there are no predictable trends (right). For additional results, see Appendix~\ref{lbl:appendix}.}
    \label{fig:generalizability_accuracy}
\end{figure}

The calligraphy dataset contained over 100,000 $64\times64$ images from 20 calligraphers (classes).
All networks were fine-tuned using Pytorch from pretrained weights (ImageNet)  until achieving high accuracy ($>$95\%) on a subset of 15 classes (in-sample). Hyperparameters: learning rate $2\mathrm{e}{-4}$, batch size 72; AdamW optimizer.
To compute $g$, we took a subset of 500 images chosen at random from each of the 5 classes not used during training or validation.
As the intermediate representation $\Phi^{i}(x)$ at each layer $i$, we used the classification token for the transformers and the vector of the mean activations of each feature map for the CNNs.
In Table \ref{table:network_comparison}, we show that \emph{generalization varies considerably across different network architectures}. In Figure \ref{fig:generalizability_accuracy}, we show that \emph{generalization varies unpredictably across layers and epochs}.


\begin{table}[t]
\caption{Generalization $g$ of classification networks for unseen and seen classes after fine-tuning.}
\begin{center}
\setlength\extrarowheight{2.5pt}
\begin{tabular}{l|ccccccc}
Network & ResNet & ViT & Swin & PViT & CvT & PoolFormer & ConvNeXtV2 \\
\hline
$g_{\text{unseen}}$ & 0.62 & 0.70 & 0.62 & 0.77 & 0.67 & {\bf 0.79} & 0.63\\
$g_{\text{seen}}$   & 0.88 & {\bf 0.95} & 0.80 & 0.93 & 0.94 & 0.91 & 0.92\\
accuracy            & 0.95 & 0.98 & 0.98 & 0.98 & 0.99 & 0.99 & 0.99 \\
\end{tabular}
\end{center}
\label{table:network_comparison}
\end{table}

\section{Conclusion}

The relationship between classification accuracy and generalization remains complex. Our experiments demonstrated the pivotal but unpredictable role architecture plays: different models maximize generalization at different layer depths; in other words, we found no consistent encoding strategy emerging after fine-tuning.
Furthermore, we found that higher accuracy does not necessarily imply higher generalizability: although all models reached high classification accuracy after fine-tuning (at least 95\%), they demonstrated differing capacities to generalize.



In our case study of calligraphy classification, we posit that the generalization index can be used as a measure of how much the network learns a general ``language'' of stroke patterns. Such a representation can then be applied to adequately represent characters from out of sample artists, as opposed to memorizing characters from in-sample artists.

More generally, we have developed a framework for quantifying the robustness of a network's representation within a domain.
Our results leave us with the following question: What capabilities might be revealed if deep learning classification performance were based on generalization rather than typical classification loss functions?

\newpage
\subsubsection*{URM Statement}
The authors acknowledge that at least one key author of this work meets the URM criteria of ICLR 2024 Tiny Papers Track.

\subsubsection*{Acknowledgments}
Supported by NIH Grant 1R01EY031059, NSF Grant 1822598.

\bibliography{iclr2023_conference_tinypaper}
\bibliographystyle{iclr2023_conference_tinypaper}

\appendix
\section{Appendix}\label{lbl:appendix}

\subsection{Alternative generalization metric}\label{lbl:app-metric}
Because NMI compares cluster assignments from k-means (section~\ref{lbl:methods}), our $g$ metric implicitly assumes that the unseen classes should form Gaussian-like distributions, thus penalizing latent embeddings in which the classes may still be reasonably separable, but more uniformly distributed. 
Therefore we tested an additional metric without this assumption, based on the traditional notion that nearest neighbors should belong to the same class. For each data point, we compute its $k$-nearest neighbors (kNN) in an embedding, with $k$
set to the number of examples in each class. We then take the mean, over all data points, of the fraction of a data point 
$x_i$'s $k$ nearest neighbors belonging to the same class as $x_i$. The results are shown in Figure~5.

\subsection{CIFAR-100 dataset}\label{lbl:app-cifar}
In addition to the calligraphy dataset, we also tested a standard natural scenes dataset, CIFAR-100. Because ImageNet-1k (the dataset used for pre-training all networks) also consists of natural scenes, we needed to ensure that there were five classes not seen during pretraining or fine-tuning. We observed that CIFAR-100 contains five classes corresponding to flower species that are not present in ImageNet-1k, namely: sunflower, tulip, orchid, poppy, and rose. Moreover, in order to have results comparable to those from the calligraphy dataset, we limited the number of CIFAR classes seen during fine-tuning to 15. Interestingly, the resulting trends were largely similar between the two datasets for all networks; see Figure~5.

\begin{figure}[h]
    \begin{center}
        \includegraphics[width=1\textwidth]{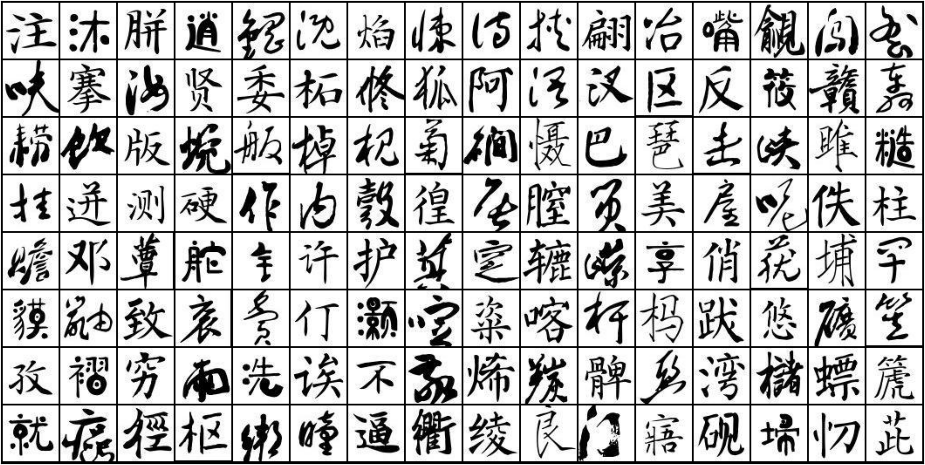} 
    \end{center}
    \caption{Subset of dataset consisting of 128 individual characters drawn by 20 different calligraphers. We invite the reader to identify characters that appear to be have been written by the same person, and to noticing the nuances of this task.
    Fine-tuned networks can easily perform this task with high accuracy for artists on which they have been trained (Table \ref{table:network_comparison}).}
    \label{fig:dataset}
\end{figure}

\begin{figure}[ht]
    \begin{center}
        \includegraphics[width=1\textwidth]{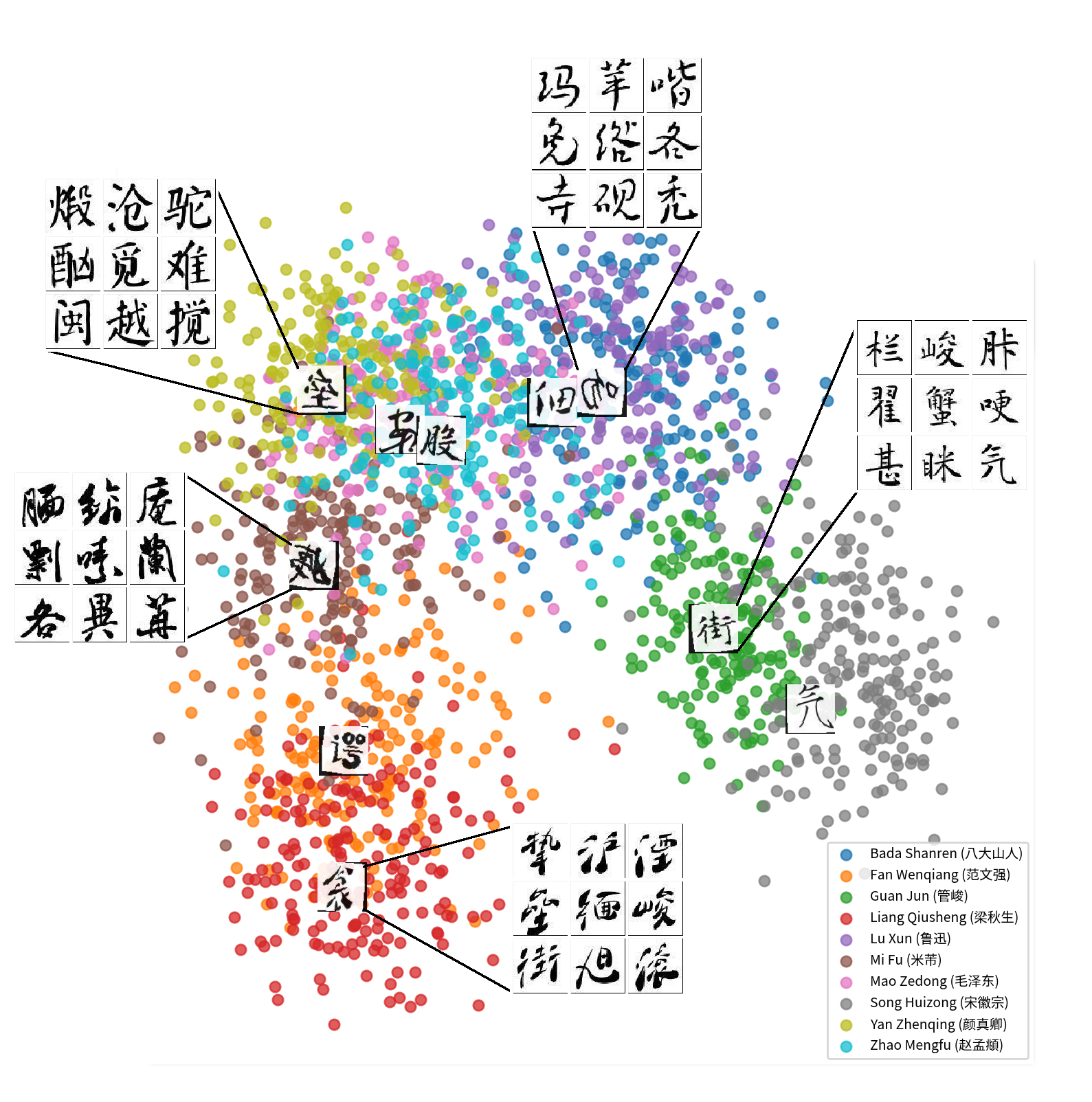}       
    \end{center}
    \caption{Embedding of 10 calligraphers (including in-sample and out-of-sample) obtained via ResNet and visualized using PCA, with representative character images for each of the ground-truth clusters shown. The images were chosen by computing the centroid of each cluster and selecting those corresponding to the 10 nearest points.}
    \label{fig:embedding_example}
\end{figure}

\begin{figure}[h]
    \begin{center}
        \includegraphics[width=1\textwidth]{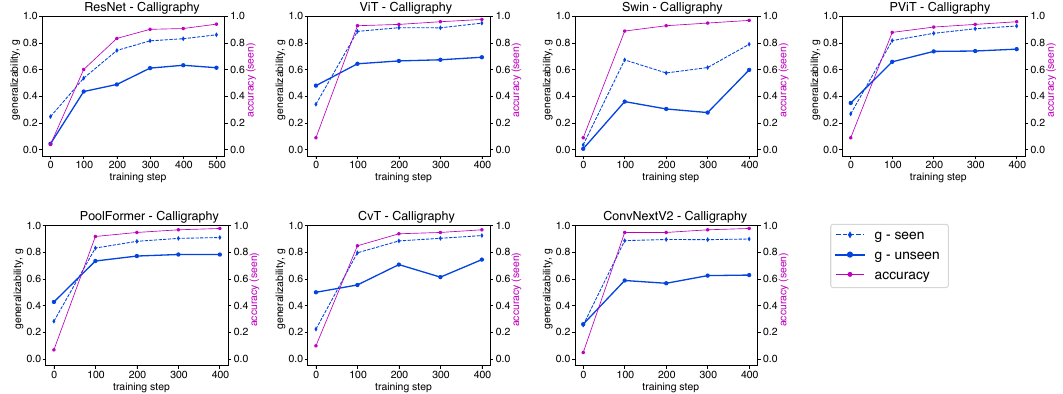}       
    \end{center}
    \caption{Generalization across fine-tuning epochs for all studied networks. We plot $g_{\mathrm{seen}}$ and test-set accuracy (using seen classes) for reference. Notice how $g_{\mathrm{seen}}$ is higher for all networks, as expected. This demonstrates that our $g$ metric is able to capture the ability of the networks to generalize to unseen classes.}
    \label{fig:accs}
\end{figure}

\begin{figure}[ht]
    \begin{center}
        \includegraphics[width=1\textwidth]{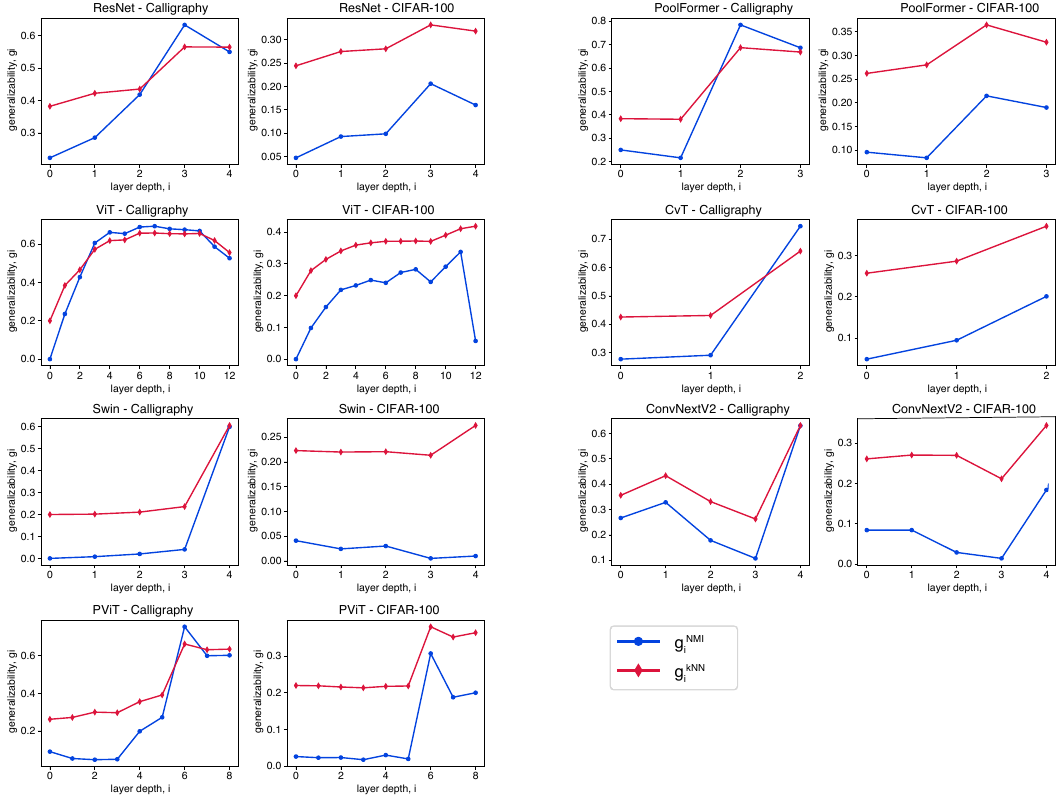}       
    \end{center}
    \caption{
The generalizability trend across layers may vary considerably from model to model. Although deeper layers tend to better separate the examples from unseen classes, $g_i$ does not increase monotonically with depth, and for most models (ResNet, ViT, PViT, and PoolFormer) the best generalization is found at intermediary layers.
This hold true for both datasets used for fine-tuning (calligraphy or CIFAR), showing that this phenomenon does not seem dataset-specific.
Moreover, similar trends were found regardless of the metric used (NMI or $k$-nearest neighbors) as demonstrated by qualitatively similar curve shapes, despite being based on completely different methods. In almost all cases, the two metrics identified the same layer as the one that best generalized to the unseen classes.
Interestingly, the $g_i$ curves are also qualitatively similar across datasets, indicating that the patterns observed are due to the network's architecture, and not the dataset.}
    \label{fig:metrics}
\end{figure}

\end{document}